\documentclass[10pt,twocolumn,letterpaper]{article}

\usepackage[pagenumbers]{cvpr} % To force page numbers, e.g. for an arXiv version

\usepackage{adjustbox}
\usepackage{colortbl} % Add this package for coloring rows
\usepackage{multirow} % For multirow and multicolumn
\usepackage{arydshln} % 点線の表を作成するためのパッケージ
\usepackage{comment}
\usepackage{pifont} % チェックマークとバツマーク用
\usepackage{siunitx}
\usepackage{tabularx}
\usepackage{xr}  % 外部参照用

\usepackage{tikz}
\usepackage{fontawesome5} % For icons
\usepackage{xcolor}

\newcommand{\name}{LoRA-TTT}
\newcommand{\namemem}{LoRA-TTT-M}
\newcommand{\namemae}{LoRA-TTT-A}
\newcommand{\nameie}{Image Encoder Tuning}

\definecolor{cvprblue}{rgb}{0.21,0.49,0.74}
\usepackage[pagebackref,breaklinks,colorlinks,allcolors=cvprblue]{hyperref}
\usepackage{amsmath,amsfonts,bm}

\def\eqref#1{equation~\ref{#1}}
\def\1{\bm{1}}

\def\vh{{\bm{h}}}

\def\vp{{\bm{p}}}

\def\vt{{\bm{t}}}

\def\vv{{\bm{v}}}

\def\vx{{\bm{x}}}

\def\mA{{\bm{A}}}
\def\mB{{\bm{B}}}

\def\mW{{\bm{W}}}

\DeclareMathAlphabet{\mathsfit}{\encodingdefault}{\sfdefault}{m}{sl}
\SetMathAlphabet{\mathsfit}{bold}{\encodingdefault}{\sfdefault}{bx}{n}

\def\sY{{\mathbb{Y}}}

\usepackage[dvipsnames]{xcolor}
\usepackage{suffix}
\newcommand\authorcomment[3]{\noindent\textsf{\textcolor{#1}{[\textbf{#2:} \textit{#3}]}}}
\WithSuffix\newcommand\authorcomment*[2]{\noindent\textcolor{#1}{\textit{#2}}}

\newcommand{\jx}[1]{\authorcomment{BurntOrange}{Jiarui}{#1}}
\WithSuffix\newcommand\jx*[1]{\authorcomment*{BurntOrange}{#1}}

\def\paperID{7613} % *** Enter the Paper ID here
\def\confName{CVPR}
\def\confYear{2025}

\title{\name: Low-Rank Test-Time Training for Vision-Language Models}

\author{
Yuto Kojima \quad Jiarui Xu \quad Xueyan Zou \quad Xiaolong Wang \\
UC San Diego \\
}

\begin{document}

\maketitle
\begin{abstract}

The rapid advancements in vision-language models (VLMs), such as CLIP, have intensified the need to address distribution shifts between training and testing datasets.
Although prior Test-Time Training (TTT) techniques for VLMs have demonstrated robust performance, they predominantly rely on tuning text prompts, a process that demands substantial computational resources and is heavily dependent on entropy-based loss.
In this paper, we propose \name, a novel TTT method that leverages Low-Rank Adaptation (LoRA), applied exclusively to the image encoder of VLMs.
By introducing LoRA and updating only its parameters during test time, our method offers a simple yet effective TTT approach, retaining the model's initial generalization capability while achieving substantial performance gains with minimal memory and runtime overhead.
Additionally, we introduce a highly efficient reconstruction loss tailored for TTT.
Our method can adapt to diverse domains by combining these two losses, without increasing memory consumption or runtime.
Extensive experiments on two benchmarks, covering 15 datasets, demonstrate that our method improves the zero-shot top-1 accuracy of CLIP-ViT-B/16 by an average of 5.79\% on the OOD benchmark and 1.36\% on the fine-grained benchmark, efficiently surpassing test-time prompt tuning, without relying on any external models or cache.

\end{abstract}
\section{Introduction}
\label{sec:intro}

\begin{figure}[ht]
    \centering
    \begin{minipage}{\linewidth}
        \centering
        \includegraphics[width=\linewidth]{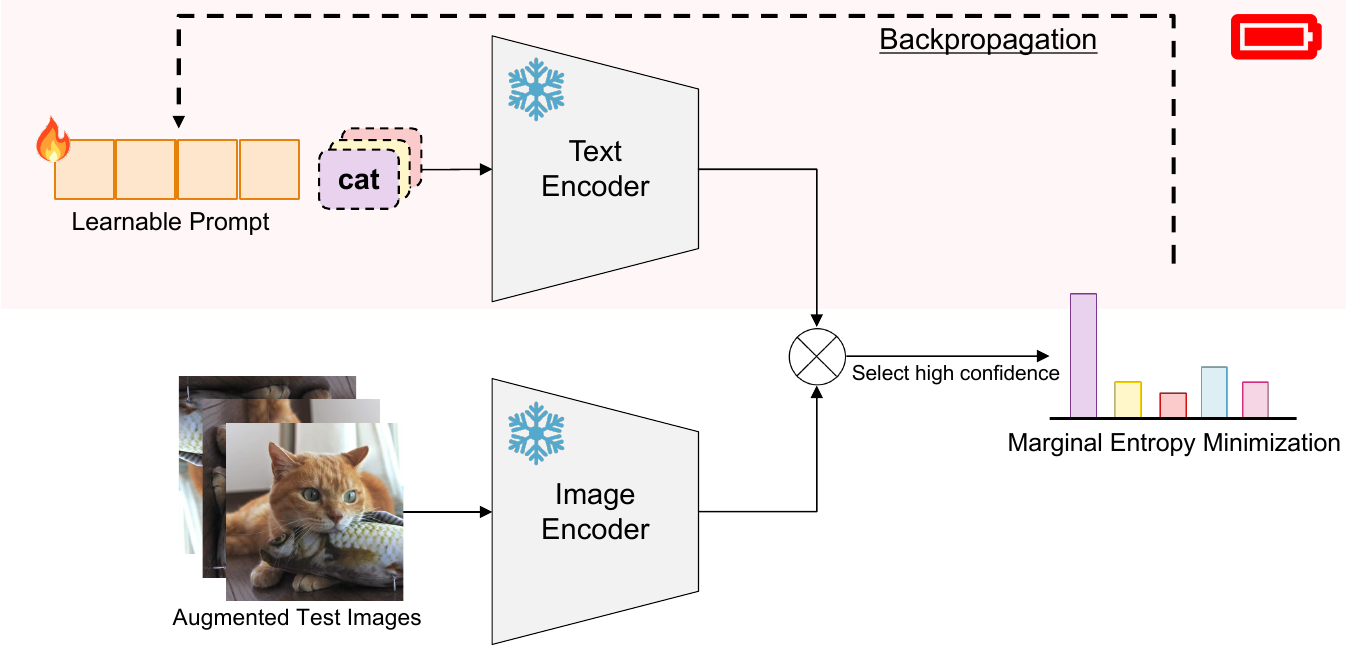}
        \subcaption{Test-time Prompt Tuning \citep{shu2022test}}
        \label{fig:tpt}
    \end{minipage}
    
    \vspace{1em} % 縦のスペース調整

    \begin{minipage}{\linewidth}
        \centering
        \includegraphics[width=\linewidth]{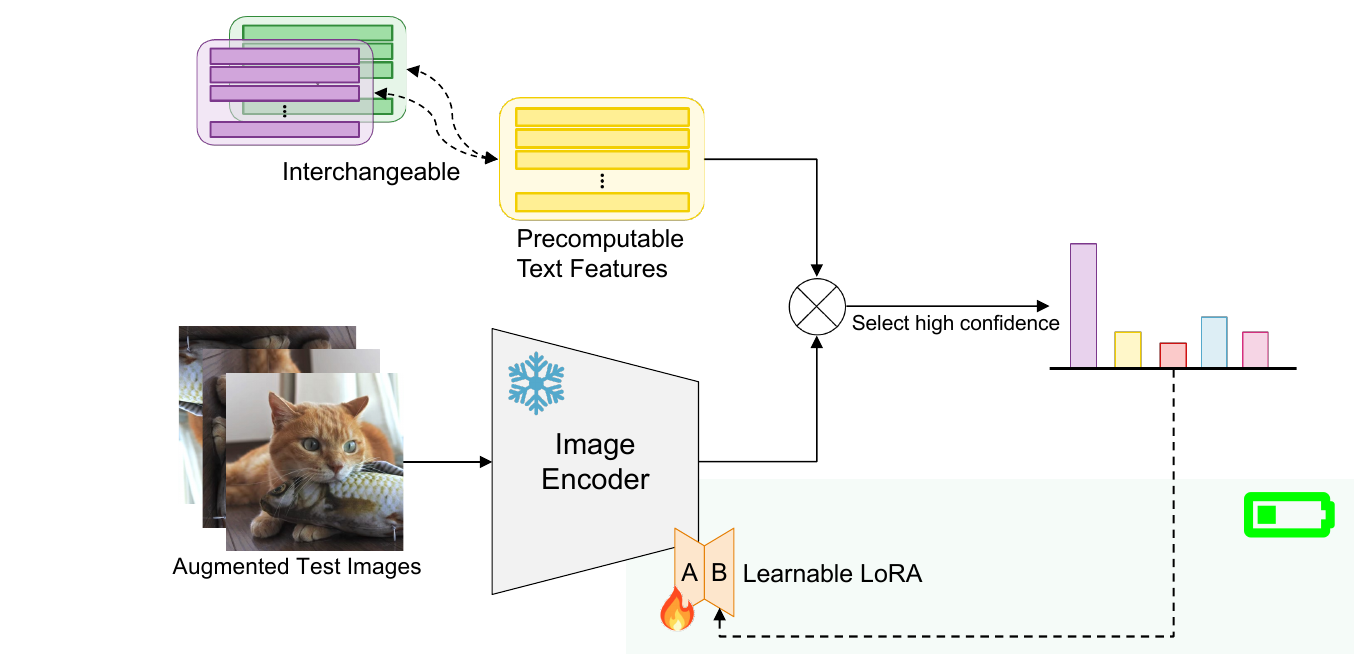}
        \subcaption{\name\ (Ours)}
        \label{fig:image_tta}
    \end{minipage}
    
    \caption{\textbf{Comparison of our proposed \name\ and Test-time Prompt Tuning (TPT).} (a) TPT optimizes the learnable text prompt via backpropagation, which results in high memory consumption, long runtime, and non-interchangeable text prompts. (b) \name\ requires only the image encoder and tunes only the LoRA parameters, demonstrating high performance while minimizing memory consumption and ensuring faster runtime. Additionally, it offers the flexibility of interchangeable text prompts.} 
    \label{fig:concept}
\end{figure}
% \jx{but change text prompt also requires re-training LoRA?}

% recent VLMs and their issue
Recent advancements in large-scale Vision-Language Models (VLMs) such as CLIP \citep{radford2021learning} and ALIGN \citep{jia2021scaling} have enabled impressive zero-shot generalization across diverse tasks, leveraging extensive pre-training on noisy, web-scale image-text pairs \citep{sharma2018conceptual, changpinyo2021conceptual, thomee2016yfcc100m}.
However, VLMs often struggle to maintain robust performance under domain shifts \citep{shu2023clipood, xiao2024any}, a common challenge in real-world applications.

The original CLIP paper \citep{radford2021learning} has demonstrated that the choice of text prompts significantly impacts zero-shot image classification performance.
As a result, the tedious work of pre-engineering text prompts for each downstream task has become increasingly important \citep{gu2023systematic,zhu2023prompt,bulat2023lasp}.
Initial works \citep{zhou2022learning,zhou2022conditional} address the burden of prompt engineering by proposing prompt fine-tuning, which directly learns text prompts using a small amount of labeled data from the target domain.
Recently, Test-Time Training (TTT) methods that adapt models during inference without requiring labeled data have gained significant attention \citep{boudiaf2022parameter,chen2022contrastive,wang2020tent}.
Although several TTT methods for VLMs aimed at achieving zero-shot generalization have been proposed \citep{shu2022test,sui2024just,yoon2024c,metzen2023autoclip}, most rely on text prompt tuning, as it is simple, requires minimal modifications, and supports black-box adaptation—an advantage for VLMs with intellectual property concerns \citep{zhang2024vision}.
% TPT issues
However, text prompt tuning often requires intensive computation \citep{karmanov2024efficient, zanella2024test}, incurs high memory consumption, and offers limited flexibility in prompt selection \citep{zanella2024test}.
For example, \cref{fig:tpt} illustrates TPT \citep{shu2022test}, a pioneering text prompt tuning method, where both the text encoder and image encoder are executed for each individual test instance, and the input text prompt — computationally intensive for backpropagation — is optimized from a fixed text prompt initialization during test time.
In practice, TTT is frequently deployed in environments with memory-constrained edge devices \citep{cai2020tinytl,song2023ecotta} and in scenarios requiring real-time data processing \citep{wang2023test,azimi2022self} or flexible prompt combinations, highlighting the practical challenges of text prompt tuning.
% Text prompt tuning is particularly inefficient for TTT, as it is often used in environments involving memory-constrained edge devices \citep{cai2020tinytl,song2023ecotta}, requires real-time processing of streaming data \citep{wang2023test,azimi2022self}, and makes it difficult to flexibly combine various prompts depending on the situation.

% Our motivation
The challenges of current TTT methods for VLMs lead to a simple question: \textit{can we improve vision representation by directly tuning the image encoder, without relying on text prompt tuning, given that downstream tasks focus on vision tasks such as image classification, segmentation, and object detection?}
Our hypothesis is that focusing on the image encoder can result in more effective and efficient adaptation in vision tasks.
However, since VLMs are pre-trained on web-scale datasets, directly tuning the parameters of the image encoder on a test instance can degrade its generalization ability and lead to catastrophic forgetting \citep{wortsman2022robust,kumar2022fine}.
Inspired by recent advancements in Parameter Efficient Fine-Tuning (PEFT) \citep{han2024parameter, xu2023parameter, ding2023parameter} for large models,
% Brief explanation of the method
we propose a \textbf{Lo}w-\textbf{Ra}nk \textbf{T}est-\textbf{T}ime \textbf{T}raining (\name) that applies Low-Rank Adaptation (LoRA) \citep{hu2021lora} to the image encoder of VLMs.
LoRA is proposed as one of the most extensively studied PEFT methods, effectively fine-tuning low-rank matrices connected in parallel to the fully connected matrices of each transformer layer, enabling the model to be tuned while preventing forgetting and minimizing memory consumption.

% explanation of the method
% \name\ offers a simple and easily implementable alternative to previous TTT methods, achieving exceptionally high performance and enhanced calibration, along with strong generalization to text prompts.
As shown in \cref{fig:image_tta}, \name\ takes a novel approach from text prompt tuning by replacing the updated parameters from text prompts with LoRA parameters using entropy loss \citep{zhang2022memo,shu2022test}.
Tuning only the LoRA parameters with a single instance during test time allows us to maintain a simple architecture and preserve the initial generalization ability, while adapting to the domain-specific features of the single instance.
\name\ focuses exclusively on vision-side parameters, eliminating the need for the text encoder during test time by precomputing and storing text features.
This reduces TTT runtime and significantly lowers memory consumption by avoiding memory-intensive text prompt tuning.
The method also generalizes well to a wide range of text prompts without relying on specific prompt designs.
% MAE utilization
In addition, focusing on the recent effectiveness of utilizing masked image modeling for TTT \citep{gandelsman2022test,wang2023test,liu2024continual}, we introduce a highly efficient reconstruction loss suited for TTT.
This loss addresses the issue of overconfidence inherent in entropy loss \citep{guo2017calibration,yoon2024c}, making it applicable in high-stakes environments \citep{wang2022medclip,liu2023clip,dorbala2022clip,khandelwal2022simple} where reliable model outputs are crucial.
By combining the entropy and reconstruction loss, our method ensures zero-shot generalization across various domains and class categories without incurring computational overhead or additional memory consumption. Our contributions can be summarized as follows:
\newline
\begin{itemize}
    \item We propose a \textbf{Lo}w-\textbf{Ra}nk \textbf{T}est-\textbf{T}ime \textbf{T}raining (\name), which applies LoRA to the image encoder of VLMs. \name\ efficiently and effectively enhances zero-shot generalization capabilities without relying on teacher models or caching, making it well-suited for real-time processing on memory-constrained edge devices or in high-stakes environments.
    \item We introduce an efficient reconstruction loss suited for TTT, which can be combined with entropy loss. It demonstrates excellent calibration performance and can be easily adapted to real-world applications.
    \item We conduct comprehensive evaluations and comparisons with existing TTT techniques for VLMs across 15 datasets from two benchmarks, achieving state-of-the-art performance by simply replacing the text prompt with LoRA for the target parameters and combining two types of losses.
\end{itemize}
\section{Related Work}
\label{sec:related_works}

\noindent\textbf{Test-Time Training (TTT)} allows models to adapt to distribution shifts between training and test data during inference through dynamic parameter updates \citep{liang2024comprehensive,wang2024search,chen2022contrastive}.
The challenges in this area lie in designing an effective test-time objective without labels and developing an efficient system suitable for real-world deployment.
For example, TENT \citep{wang2020tent} tunes batch normalization statistics at test time using entropy loss; however, this approach requires batch processing rather than instance-level processing, making it challenging to handle sequential data in real-time.
Sun \etal~\cite{sun2020test} and Gandelsman \etal~\cite{gandelsman2022test} update the image encoder by introducing auxiliary tasks and applying self-supervision; however, these methods require fine-tuning the model with auxiliary tasks beforehand for TTT.
% For VLMs, methods that primarily tune text prompts \citep{zhou2022learning,zhou2022conditional}, visual prompts \citep{jia2022visual,bahng2022exploring}, or both \citep{khattak2023maple} have been proposed to address distribution shift, owing to their simplicity and effectiveness.
For VLMs, TPT \citep{shu2022test} focuses on optimizing a text prompt at test time, valued for its simplicity and effectiveness.
It demonstrates that augmenting a single test instance and calculating marginal entropy minimization \citep{zhang2022memo} serves as an effective loss for VLMs.
DiffTPT \citep{feng2023diverse} utilizes stable diffusion to enhance data augmentation quality, while C-TPT \citep{yoon2024c} is a technique that calibrates TPT to improve reliability.
RLCF \citep{zhao2023test} tunes the image encoder and demonstrates that CLIP-ViT-B can achieve performance comparable to CLIP-ViT-L but requires CLIP-ViT-L as a feedback source, which poses challenges in memory-constrained environments.
Our method requires no external resources and remains feasible even in closed memory-constrained environments such as edge devices.
\newline

\noindent\textbf{Application of Low-rank adaptation (LoRA)} aims to achieve efficient fine-tuning of large models with vast numbers of parameters in memory-constrained environments by introducing trainable low-rank matrices into each layer of the Transformer architecture, allowing the pre-trained parameters to remain frozen \citep{hu2021lora,han2024parameter,xin2024parameter}.
MeLo \citep{zhu2024melo} demonstrates that applying LoRA to vision transformers  (ViT) for downstream medical image diagnosis achieves comparable performance to fully fine-tuned ViT models while significantly reducing memory consumption.
CLIP-LoRA \citep{zanella2024low} demonstrate significant performance improvements in few-shot learning by applying LoRA to the vision encoder of CLIP.
However, CLIP-LoRA requires a few labeled samples from the target downstream task.
To the best of our knowledge, no approach has applied LoRA for TTT in VLMs. 

\section{Method}

\begin{figure*}[!htbp]
\centering
\includegraphics[width=1.0\textwidth]{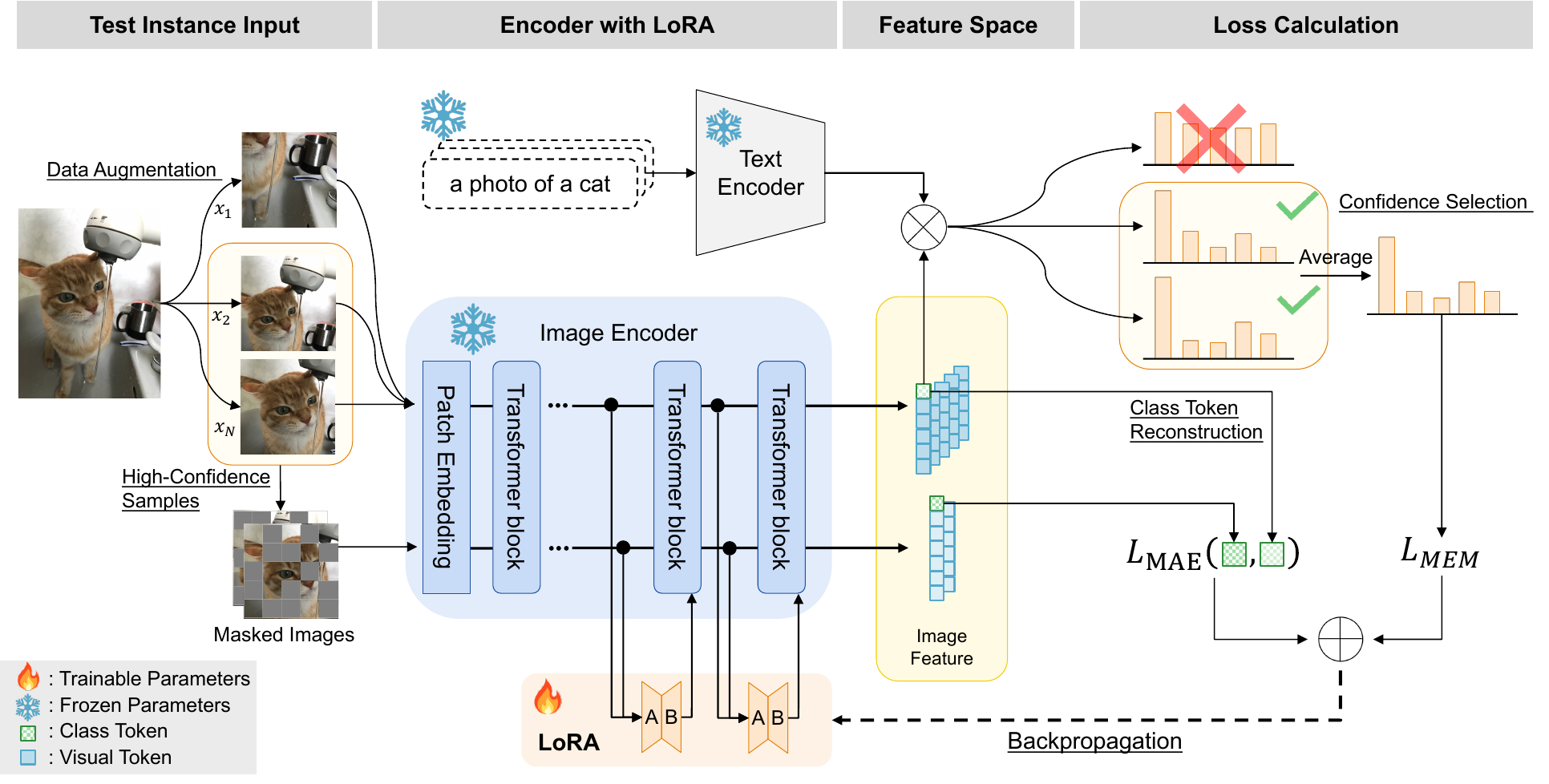}
\caption{\textbf{\name} for zero-shot image classification. Our method applies LoRA to the layers of the image encoder in VLMs. \name\ updates the LoRA parameters using MEM loss and MAE loss, calculated from the top 10\% of high-confidence augmented views. This approach allows adaptation to domain shifts with low memory consumption while maintaining generalization ability.}
\label{fig:method}
\end{figure*}

\subsection{Preliminaries}

\textbf{Contrastive Language-Image Pre-training (CLIP).}\hspace{5mm}
Pre-training VLMs is typically conducted with specific vision-language objectives that facilitate learning image-text correspondences from extensive image and text datasets \citep{radford2021learning,yu2022coca,yao2021filip}.
CLIP \citep{radford2021learning} consists of an image encoder $g$ and a text encoder $f$, pre-trained using a contrastive loss to align the embeddings from both encoders in a shared feature space.
After pre-training, CLIP exhibits strong zero-shot capabilities across various downstream tasks \citep{radford2021learning,minderer2022simple,wang2021actionclip,li2022language}.
For zero-shot image classification with CLIP, input text prompts are created by inserting each of the K-class category labels $\sY = {y_1, y_2, \dots, y_K}$ into a prompt prefix $\vp$ (e.g., $\vp_i$ = "a photo of a $y_i$").
% it creates input text prompts by adding each of the K-class category labels $\sY = \{y_1, y_2, \dots, y_K\}$ into a prompt prefix $\vp$ (\eg, $\vp_i$ =``a photo of a $y_i$'').
The image $X$ and the text input $\vp_i$ representing the $i$-th class are encoded into $\vv=g(X)$ and $\vt_i=f(\vp_i)$ by their respective encoders. The classification score for the $i$-th class of image $X$ is then calculated by
\begin{equation}
p(y_i\mid X) = \frac{\exp(\text{cos}(\vt_i \cdot \vv) / \tau)}{\sum_{i=1}^{K} \exp(\text{cos}(\vt_i \cdot \vv) / \tau)}
\end{equation}
in zero-shot fashion, where $\tau$ is the temperature parameter and $\text{cos}(\cdot, \cdot)$ represents the cosine similarity.
\newline

\noindent\textbf{TPT} \citep{shu2022test} makes the text prompt prefix trainable $\vp \in \mathbb{R}^{L \times D}$ instead of using pre-determined, hand-crafted prompt, where $L$ represents the number of tokens and $D$ is the embedding size of the text encoder.
TPT optimizes the learnable text prompt using backpropagation with Marginal Entropy Minimization (MEM) loss \citep{zhang2022memo} from each test instance.
A test image instance undergoes data augmentation $N$ times $\{X_n\}_{n=1}^{N}$, and the top $N_p$ samples with the highest confidence are selected to calculate the MEM loss, filtering out noisy augmented views.
The entropy of the predicted probability distribution over the K-classes is calculated by
\begin{equation}
% \mathcal{L}_{\text{MEM}} = H\left(\bar{p}(\cdot \mid X)\right) = -\sum_{i=1}^{K} \bar{p}(y_i \mid X) \log(\bar{p}(y_i \mid X)),
\mathcal{L}_{\text{MEM}} = -\sum_{i=1}^{K} \bar{p}(y_i \mid X) \log(\bar{p}(y_i \mid X)),
\end{equation}
where
$\bar{p}(\cdot \mid X) = \frac{1}{N_p} \sum_{n=1}^{N_p} p(\cdot \mid X_n)$.
MEM loss has become the de facto standard in modern TTT for VLMs. \citep{farina2024frustratingly}.
\newline

\noindent\textbf{LoRA} \citep{hu2021lora} represents the incremental adjustment of pre-trained weights by injecting two small matrices, based on the concept of the intrinsic rank of downstream domain shifts \citep{aghajanyan2020intrinsic}.
Given that $\mW_0 \in \mathbb{R}^{d_1 \times d_2}$ is a pre-trained weight matrix in the network, with $\vx$ as the input and $\displaystyle \vh$ as the hidden state, the forward pass after applying LoRA is given by:
\begin{equation}
\label{eq:lora}
\vh = \mW_0 \vx + \Delta \mW \vx =  \mW_0 \vx + \gamma\mB\mA\vx,
\end{equation}
where $\mA \in \mathbb{R}^{r \times d_2}$ and $\mB \in \mathbb{R}^{d_1 \times r}$ are the two small matrices introduced by LoRA.
$\gamma$ is a constant scale hyper-parameter that determines the contribution of LoRA in the subsequent layers. The intrinsic rank $r$ is typically much smaller than $d_1$ and $d_2$. Therefore, by freezing the pre-trained matrix and updating only the parameters of the $\mA$ and $\mB$ matrices during fine-tuning, the number of parameters that need to be updated can be significantly reduced. In the original LoRA paper, LoRA is applied to the key, query, value, and output matrices on the attention matrices of transformer-based models. 

\subsection{\name}

\textbf{Application of LoRA for TTT.}\hspace{5mm}
Focusing on the current mainstream method, TPT, our approach shifts the target of parameter updates from the text prompt to the image encoder, while leveraging the MEM loss.
However, directly updating the entire image encoder is expected to result in excessive memory consumption and domain-specific behaviors that lose the out-of-distribution generalization and robustness of foundation models \citep{wortsman2022robust, kumar2022fine}. % this sentence comes from TPT paper
As shown in \cref{fig:method}, inspired by the effectiveness of LoRA in the large language model field, we apply LoRA to layers of the image encoder in VLMs, updating only the LoRA parameters during test time.
The original LoRA paper shows that the change in weights during model adaptation has a low intrinsic rank and we hypothesize that LoRA can adapt to unique domain-specific features of each instance even with unlabeled data similar to fine-tuning in downstream tasks.
By applying LoRA, we adjust only a small number of parameters without altering the original well pre-trained weights in VLMs.
This approach allows individual adaptation to each test instance while preserving the strong zero-shot capability of the original VLMs and reducing memory consumption during backpropagation.
Furthermore, \name\ applies LoRA exclusively to the vision encoder of VLMs, rendering the text encoder unnecessary during TTT.
This allows the model to be tuned independently of specific text prompts.
Following the approach of Episodic TTT \citep{wang2020tent,shu2022test,zhao2023test}, we update parameters using only a single test instance and reset them afterward, ensuring a certain level of robustness against sequence data.
\newline

\noindent\textbf{Masked Image Reconstruction for VLMs.}\hspace{5mm}
\name\ is not limited by the type of loss function.
We are able to leverage a self-supervised reconstruction loss based on masked autoencoders (MAE) \citep{he2022masked, gandelsman2022test}, owing to the benefits of parameterizing the image encoder.
\name\ differs from conventional TTT methods based on MAE \citep{gandelsman2022test,wang2023test} and offers an efficient solution suitable for TTT, as it requires neither an image decoder nor fine-tuning of the model prior to TTT.
% differs from conventional TTT methods based on MAE \citep{gandelsman2022test,wang2023test} and offers an efficient solution suitable for TTT, as it does not require an image decoder. % ?
\name\ takes both augmented images and their randomly masked versions as input to the image encoder and calculates the mean squared error of only the encoded class tokens as the loss. These images are selected from the top 10\% of views with the highest confidence, similar to TPT.
We optimize the following loss:
\begin{equation}
\mathcal{L}_{\text{MAE}} = \text{MSE}(g(X)_{\text{cls}}, g(\text{mask}(X))_{\text{cls}}),
\end{equation}
where $\text{MSE}(\cdot, \cdot)$ represents the mean squared error between the encoded class tokens of the masked and unmasked images, $\text{mask}()$ randomly masks out majority of the input image patches (\eg, 50\%). 
This loss encourages the model to reconstruct the original global features by leveraging the remaining visual clues, enhancing visual understanding to better support downstream tasks.
The total loss can be expressed as
$
\mathcal{L} = \lambda_1\mathcal{L}_{\text{MEM}} + \lambda_2\mathcal{L}_{\text{MAE}},
$ where $\lambda_1$ and $\lambda_2$ are coefficients that balance the two losses.
\section{Experiments}
\label{sec:experiments}

\begin{table*}[ht]
\centering
\caption{\textbf{Top1 accuracy of zero-shot image classification on the OOD benchmark} when using the default hard prompt.
The results of CoCoOp are obtained from the TPT paper, while others are reproduced with our code. 
The best results under zero-shot conditions are highlighted in \textbf{bold}. Performance improvements over the zero-shot CLIP-ViT-B/16 are indicated with an upward blue arrow {\textcolor{blue}{($\uparrow$blue)}} and a downward red arrow {\textcolor{red}{($\downarrow$red)}}.}
\label{tab:imagenets_result}
\begin{adjustbox}{width=0.9\textwidth}
\begin{tabular}{lccccccc}
\toprule
\rowcolor{gray!10} \textbf{Method} & \textbf{ImageNet} & \textbf{ImageNet-A} & \textbf{ImageNet-V2} & \textbf{ImageNet-R} & \textbf{ImageNet-Sketch} & \textbf{Average} & \textbf{OOD Avg.} \\
\midrule
CLIP-ViT-B/16 & 66.71 & 47.80 & 60.63 & 73.99 & 46.15 & 59.06 & 57.14 \\
\midrule
CoOp \citep{zhou2022learning} & 71.75 & 50.13 & 64.51 & 75.28 & 47.92 & 61.92 & 59.46 \\
CoCoOp \citep{zhou2022conditional} & 71.02 & 50.63 & 64.07 & 76.18 & 48.75 & 62.13 & 59.91 \\
\midrule
TPT \citep{shu2022test} & 69.02 & 54.73 & 63.70 & 77.15 & 47.99 & 62.52 & 60.89 \\ 
C-TPT \citep{yoon2024c} & 68.50 & 51.60 & 62.70 & 76.00 & 47.90 & 61.34 & 59.55 \\
MTA \citep{zanella2024test} & 69.23 & 56.87 & 63.67 & 76.88 & 48.54 & 63.04 & 61.49 \\
\midrule
\nameie & 64.26 & 56.31 & 59.70 & 75.89 & 47.65 & 60.76 & 59.89 \\ 
\rowcolor{blue!10} \textbf{\namemem} (Ours) & 69.21\textsubscript{\textcolor{blue}{($\uparrow$2.49)}} & \textbf{60.57}\textsubscript{\textcolor{blue}{($\uparrow$12.77)}} & 64.28\textsubscript{\textcolor{blue}{($\uparrow$3.65)}} & 77.53\textsubscript{\textcolor{blue}{($\uparrow$3.54)}} & 48.73\textsubscript{\textcolor{blue}{($\uparrow$2.57)}} & 64.06\textsubscript{\textcolor{blue}{($\uparrow$5.01)}} & 62.78\textsubscript{\textcolor{blue}{($\uparrow$5.64)}} \\
\rowcolor{blue!5} \textbf{\namemae} (Ours) & 66.27\textsubscript{\textcolor{red}{($\downarrow$0.45)}} & 52.55\textsubscript{\textcolor{blue}{($\uparrow$4.75)}} & 60.87\textsubscript{\textcolor{blue}{($\uparrow$0.24)}} & 75.57\textsubscript{\textcolor{blue}{($\uparrow$1.58)}} & 47.01\textsubscript{\textcolor{blue}{($\uparrow$0.85)}} & 60.45\textsubscript{\textcolor{blue}{($\uparrow$1.39)}} & 59.00\textsubscript{\textcolor{blue}{($\uparrow$1.86)}} \\
\rowcolor{blue!15} \textbf{\name} (Ours) & \textbf{69.40}\textsubscript{\textcolor{blue}{($\uparrow$2.68)}} & 60.52\textsubscript{\textcolor{blue}{($\uparrow$12.72)}} & \textbf{64.43}\textsubscript{\textcolor{blue}{($\uparrow$3.80)}} & \textbf{77.84}\textsubscript{\textcolor{blue}{($\uparrow$3.85)}} & \textbf{48.94}\textsubscript{\textcolor{blue}{($\uparrow$2.79)}} & \textbf{64.23}\textsubscript{\textcolor{blue}{($\uparrow$5.17)}} & \textbf{62.93}\textsubscript{\textcolor{blue}{($\uparrow$5.79)}} \\
\bottomrule
\end{tabular}
\end{adjustbox}
\end{table*}

This section reports benchmark results for zero-shot image classification, comparing our proposed approach with previous methods.
Following prior work \citep{shu2022test,feng2023diverse,karmanov2024efficient},  we evaluate out-of-distribution (OOD) performance on 4 datasets derived from ImageNet and fine-grained (FG) classification on 10 datasets spanning various categories.

\subsection{Experimental setup}
\label{dataset}

\textbf{Datasets.}\hspace{5mm}
For the OOD benchmark, we use ImageNet \citep{deng2009imagenet} and its variants including ImageNet-A \citep{hendrycks2021natural}, ImageNet-V2 \citep{recht2019imagenet}, ImageNet-R \citep{hendrycks2021many}, and ImageNet-Sketch \citep{wang2019learning}, to evaluate robustness against 4 out-of-distribution patterns derived from ImageNet.
Additionally, we evaluate 10 different datasets for the fine-grained benchmark. These datasets encompass a wide range of categories, including plants and animals (Flower102 \citep{nilsback2008automated}, OxfordPets \citep{parkhi2012cats}), scene recognition (SUN397 \citep{xiao2010sun}), textures (DTD \citep{cimpoi2014describing}), food (Food101 \citep{bossard2014food}), transportation (StanfordCars \citep{krause20133d}, Aircraft \citep{maji2013fine}), human actions (UCF101 \citep{soomro2012ucf101}), satellite images (EuroSAT \citep{helber2019eurosat}), and general objects (Caltech101 \citep{li2022caltech}). This benchmark assesses the applicability of our TTT method across a diverse range of categories.
% The details of each dataset are provided in Appendix \ref{sec:datasets}.
\newline

\noindent\textbf{Baselines.}\hspace{5mm}
We compare our method with the baseline CLIP-ViT-B/16 and few-shot learning methods — CoOp \citep{zhou2022learning}, CoCoOp \citep{zhou2022conditional} — as well as existing test-time prompt tuning methods — TPT \citep{shu2022test} and C-TPT \citep{yoon2024c}.
Additionally, we compare \nameie, which tunes the image encoder parameters without relying on LoRA, and MTA \citep{zanella2024test}, which operates without backpropagation.
For a fair comparison, we focus on methods that do not rely on external models or cached data.
\newline

\noindent\textbf{Implementation details.}\hspace{5mm}
We adopt the pre-trained CLIP-ViT-B/16 as the common backbone architecture.
In text prompt tuning methods, the number of trainable text tokens is set to 4, with initial weights based on the prompt ``a photo of a''.
We prepare three versions as precomputed text prompts: the default hard prompt ``a photo of a'' to match the initialization commonly used in text prompt tuning, an ensemble of 80 different hand-crafted prompts \cite{radford2021learning}, and CoOp \citep{zhou2022learning}.
The weights of CoOp are pre-trained on ImageNet with 16 shots and 4 tokens.
LoRA is applied exclusively to the transformer architecture in layers 11 and 12 of the image encoder with a rank of 16 targeting the key, query, value, and output matrices.
The LoRA scale $\gamma$ is set to 12 for the OOD benchmark and 2 for the fine-grained benchmark.
Matrix A of LoRA is initialized using Kaiming-uniform \citep{he2015delving}, while matrix B is initialized to zero.
For the test-time loss variants of our method, \textbf{\namemem} uses the MEM loss only, \textbf{\namemae} uses the MAE loss only, and \textbf{\name} combines the two losses with the weights set to $\lambda_1 = 1$ and $\lambda_2 = 16$.
We optimize the LoRA parameters in a single step, using the AdamW optimizer \citep{loshchilov2017decoupled} with a learning rate of 0.001 and a weight decay value of 0.2.
For \nameie, we use the same optimizer and loss settings as in \textbf{\name} for a fair comparison, directly tuning the parameters of the key, query, value, and output matrices in the transformer layers 11 and 12 of the image encoder.
Data augmentation follows TPT by expanding a single instance into a 64-batch using random resized crops, including the original instance.
Additionally, the top 10\% of high-confidence samples from the batch of 64 are selected to compute the test losses and filter out noisy views.
All the experiments are conducted using a single NVIDIA RTX 3090 GPU with 24GB of memory.

\subsection{Results}
\begin{table*}[h]
\centering
\caption{\textbf{Top-1 accuracy of zero-shot image classification on the fine-grained benchmark} using the default hard prompt. CoCoOp is from the TPT paper, while others are reproduced with our code. The best results under zero-shot conditions are in \textbf{bold}. Performance improvements over zero-shot CLIP-ViT-B/16 are indicated with an upward blue arrow {\textcolor{blue}{($\uparrow$blue)}} and a downward red arrow {\textcolor{red}{($\downarrow$red)}}.}
\label{tab:fg_result}
\begin{adjustbox}{width=\textwidth}
\begin{tabular}{lccccccccccc}
\toprule
\rowcolor{gray!10} \textbf{Method}  & \textbf{Flower102} & \textbf{DTD} & \textbf{Pets} & \textbf{Cars} & \textbf{UCF101} & \textbf{Caltech} & \textbf{Food101} & \textbf{SUN397} & \textbf{Aircraft} & \textbf{EuroSAT} & \textbf{FG Avg.} \\
\midrule
CLIP-ViT-B/16 & 67.40 & 44.39 & 88.25 & 65.51 & 65.24 & 93.31 & 83.64 & 62.56 & 23.91 & 42.22 & 63.64 \\
\midrule
CoOp \citep{zhou2022learning} & 68.30 & 42.34 & 89.35 & 63.30 & 67.19 & 92.85 & 83.72 & 64.53 & 19.96 & 40.19 & 63.17 \\
CoCoOp \citep{zhou2022conditional} & 70.85 & 45.45 & 90.46 & 64.90 & 68.44 & 93.79 & 83.97 & 66.89 & 22.29 & 39.23 & 64.63 \\
\midrule
TPT \citep{shu2022test} & 68.98 & 45.92 & 87.27 & 67.02 & \textbf{68.99} & 93.55 & \textbf{85.00} & \textbf{65.11} & 23.76 & 43.44 & 64.91 \\ 
C-TPT \citep{yoon2024c} & \textbf{69.67} & 44.80 & 88.47 & 65.97 & 65.27 & 93.35 & 83.23 & 64.28 & 23.97 & 42.21 & 64.12 \\
MTA \citep{zanella2024test} & 68.29 & 45.33 & 88.17 & \textbf{68.08} & 68.07 & \textbf{94.12} & 84.88 & 64.72 & 25.38 & 40.91 & 64.79 \\
\midrule
\nameie & 59.03 & 42.91 & 83.81 & 66.60 & 67.57 & 88.64 & 82.11 & 62.71 & 23.67 & 36.67 & 61.37 \\

\rowcolor{blue!10}\textbf{\namemem} (Ours) & 67.60\textsubscript{\textcolor{blue}{($\uparrow$0.20)}} & \textbf{46.04}\textsubscript{\textcolor{blue}{($\uparrow$1.65)}} & 87.11\textsubscript{\textcolor{red}{($\downarrow$1.14)}} & 67.81\textsubscript{\textcolor{blue}{($\uparrow$2.30)}} & 68.38\textsubscript{\textcolor{blue}{($\uparrow$3.15)}} & 93.59\textsubscript{\textcolor{blue}{($\uparrow$0.28)}} & 84.83\textsubscript{\textcolor{blue}{($\uparrow$1.19)}} & 64.61\textsubscript{\textcolor{blue}{($\uparrow$2.05)}} & 25.68\textsubscript{\textcolor{blue}{($\uparrow$1.77)}} & 39.27\textsubscript{\textcolor{red}{($\downarrow$2.95)}} & 64.49\textsubscript{\textcolor{blue}{($\uparrow$0.85)}} \\

\rowcolor{blue!5}\textbf{\namemae} (Ours) & 68.33\textsubscript{\textcolor{blue}{($\uparrow$0.93)}} & 45.21\textsubscript{\textcolor{blue}{($\uparrow$0.83)}} & \textbf{88.72}\textsubscript{\textcolor{blue}{($\uparrow$0.46)}} & 66.94\textsubscript{\textcolor{blue}{($\uparrow$1.43)}} & 66.35\textsubscript{\textcolor{blue}{($\uparrow$1.11)}} & 93.71\textsubscript{\textcolor{blue}{($\uparrow$0.41)}} & 84.39\textsubscript{\textcolor{blue}{($\uparrow$0.75)}} & 63.63\textsubscript{\textcolor{blue}{($\uparrow$1.07)}} & 25.38\textsubscript{\textcolor{blue}{($\uparrow$1.47)}} & \textbf{44.52}\textsubscript{\textcolor{blue}{($\uparrow$2.30)}} & 64.72\textsubscript{\textcolor{blue}{($\uparrow$1.08)}} \\

\rowcolor{blue!15}\textbf{\name} (Ours) & 67.88\textsubscript{\textcolor{blue}{($\uparrow$0.49)}} & 45.86\textsubscript{\textcolor{blue}{($\uparrow$1.48)}} & 87.63\textsubscript{\textcolor{red}{($\downarrow$0.63)}} & 67.72\textsubscript{\textcolor{blue}{($\uparrow$2.20)}} & 68.38\textsubscript{\textcolor{blue}{($\uparrow$3.15)}} & 93.83\textsubscript{\textcolor{blue}{($\uparrow$0.53)}} & 84.99\textsubscript{\textcolor{blue}{($\uparrow$1.35)}} & 64.59\textsubscript{\textcolor{blue}{($\uparrow$2.03)}} & \textbf{25.92}\textsubscript{\textcolor{blue}{($\uparrow$2.01)}} & 43.23\textsubscript{\textcolor{blue}{($\uparrow$1.01)}} & \textbf{65.00}\textsubscript{\textcolor{blue}{($\uparrow$1.36)}} \\
\bottomrule
\end{tabular}
\end{adjustbox}
\end{table*}

\begin{table}[t]
\centering
\caption{\textbf{Generalization in prompts.} Performance differences from each single hard prompt are indicated with an upward blue arrow {\textcolor{blue}{($\uparrow$blue)}} and a downward red arrow {\textcolor{red}{($\downarrow$red)}}.}
\label{tab:prompt}
\begin{adjustbox}{width=\linewidth}
\begin{tabular}{lccc}
\toprule
\rowcolor{gray!10} \textbf{Method} & \textbf{ImageNet} & \textbf{OOD Average} & \textbf{FG Average} \\
\midrule
CLIP-ViT-B/16 + Ensemble & 68.31\textsubscript{\textcolor{blue}{($\uparrow$1.59)}} & 59.52\textsubscript{\textcolor{blue}{($\uparrow$2.38)}} & 64.68\textsubscript{\textcolor{blue}{($\uparrow$1.04)}} \\
TPT + Ensemble & 67.21\textsubscript{\textcolor{red}{($\downarrow$1.81)}} & 59.93\textsubscript{\textcolor{red}{($\downarrow$0.96)}} & 63.37\textsubscript{\textcolor{red}{($\downarrow$1.54)}} \\ 
\rowcolor{blue!10}\textbf{\name + Ensemble} (Ours) & \textbf{70.67}\textsubscript{\textcolor{blue}{($\uparrow$1.27)}} & \textbf{64.99}\textsubscript{\textcolor{blue}{($\uparrow$2.06)}} & \textbf{65.95}\textsubscript{\textcolor{blue}{($\uparrow$0.95)}} \\
\midrule
CLIP-ViT-B/16 + CoOp & 71.75\textsubscript{\textcolor{blue}{($\uparrow$5.03)}} & 59.46\textsubscript{\textcolor{blue}{($\uparrow$2.32)}} & 63.17\textsubscript{\textcolor{red}{($\downarrow$0.47)}}\\
TPT + CoOp & 73.63\textsubscript{\textcolor{blue}{($\uparrow$4.62)}} & 62.98\textsubscript{\textcolor{blue}{($\uparrow$2.09)}} & 63.95\textsubscript{\textcolor{red}{($\downarrow$0.95)}}\\
\rowcolor{blue!10}\textbf{\name + CoOp} (Ours) & \textbf{74.03}\textsubscript{\textcolor{blue}{($\uparrow$4.63)}} & \textbf{64.35}\textsubscript{\textcolor{blue}{($\uparrow$1.42)}} & \textbf{64.00}\textsubscript{\textcolor{red}{($\downarrow$1.01)}}\\
\bottomrule
\end{tabular}
\end{adjustbox}
\end{table}

\textbf{Zero-shot classification.}\hspace{5mm}
In \cref{tab:imagenets_result} and \cref{tab:fg_result}, \name\ enhances the zero-shot generalization capacity of CLIP-ViT-B/16, outperforming both few-shot learning and text prompt tuning methods on average across the two benchmark evaluations. Notably, it significantly surpasses existing methods on the OOD benchmark, demonstrating that it is a highly effective approach for adapting to various domains.
% \name\ improves the zero-shot generalization capacity of CLIP-ViT-B/16, outperforming both the few-shot learning and the text prompt tuning methods on average across the two benchmark evaluations, demonstrating that it is a highly versatile approach capable of adapting to various domains.
\name\ achieves state-of-the-art performance among methods that do not rely on domain knowledge, additional models, or cache.
Despite \nameie\ failing to adapt effectively to the fine-grained benchmark, \name\ consistently delivers significantly better results, showing improvements of 3.04\% in OOD and 3.63\% in fine-grained averages compared to \nameie.
This confirms that LoRA tuning is effective for adapting to domain gaps in VLMs, helping to prevent forgetting and making it a suitable approach for TTT.

In the comparison of the two test-time losses, \namemem\ and \namemae, both methods show performance improvements over the baseline.
In \cref{tab:fg_result}, \namemae\ (\ie, MAE loss) achieves comparable or better results than \namemem\ (\ie, MEM loss) across a wide range of category domains in the fine-grained benchmark, particularly demonstrating the high versatility of MAE as a test-time loss in VLMs.
Although our use of MAE loss involves only an image encoder without a decoder, it demonstrates that restoring global features of masked images contributes to understanding image context in VLMs.
Combining the two losses further enhances both versatility and performance, surpassing TPT in the fine-grained benchmark.
For example, a comparison of TPT with \namemem\ on EuroSAT highlights the importance of tuning the text prompt when using only the MEM loss, but the combination with the MAE loss helps overcome this weakness.
\newline

\noindent\textbf{Generalization in prompts.}\hspace{5mm}
\cref{tab:prompt} shows the generalization performance when using either the ensemble of text prompts or CoOp.
TPT is designed to initialize a single hard prompt, making it not straightforward to combine with the ensemble of prompts.
In our implementation of TPT, we initialize the prompt using the average of embeddings from the text prompts.
By combining our method with the ensemble, we achieve performance improvements comparable to those of CLIP-ViT-B/16 with the ensemble, significantly surpassing the state-of-the-art in text prompt tuning while maintaining zero-shot conditions.
Even when combined with CoOp, a few-shot learning approach, \name\ shows a similar trend to CLIP-ViT-B/16 + CoOp, indicating that our method improves performance independently of the text prompt.
Our method focuses solely on tuning the parameters within the image encoder, allowing for greater flexibility in designing text prompts, an advantage in practical applications where the ability to freely choose prompts is beneficial \citep{gu2023systematic}.
\newline

\begin{figure}[t]
  \centering
    \begin{minipage}{0.32\linewidth}
        \centering
        \includegraphics[width=\linewidth]{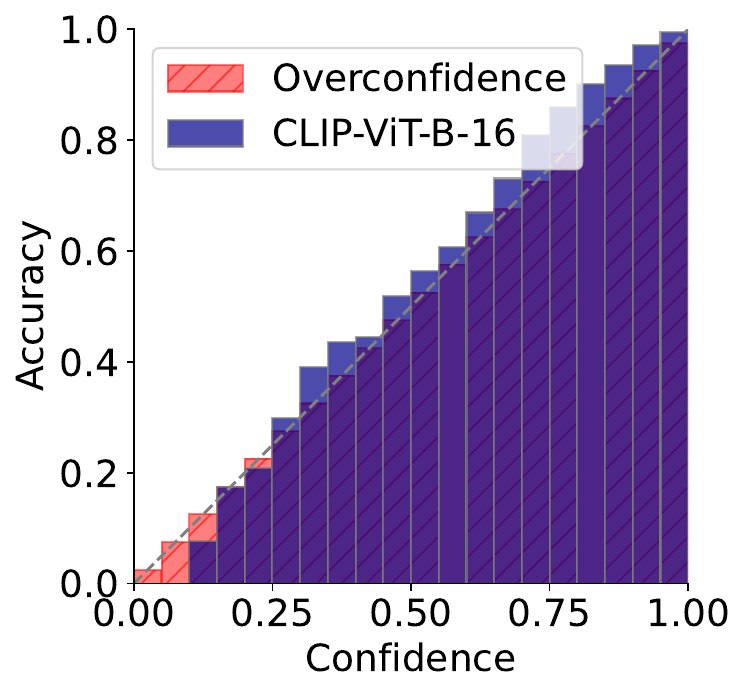}
        \subcaption{CLIP-ViT-B/16}
        \label{fig:calib_baseline}
    \end{minipage}
    \hfill
    \begin{minipage}{0.32\linewidth}
        \centering
        \includegraphics[width=\linewidth]{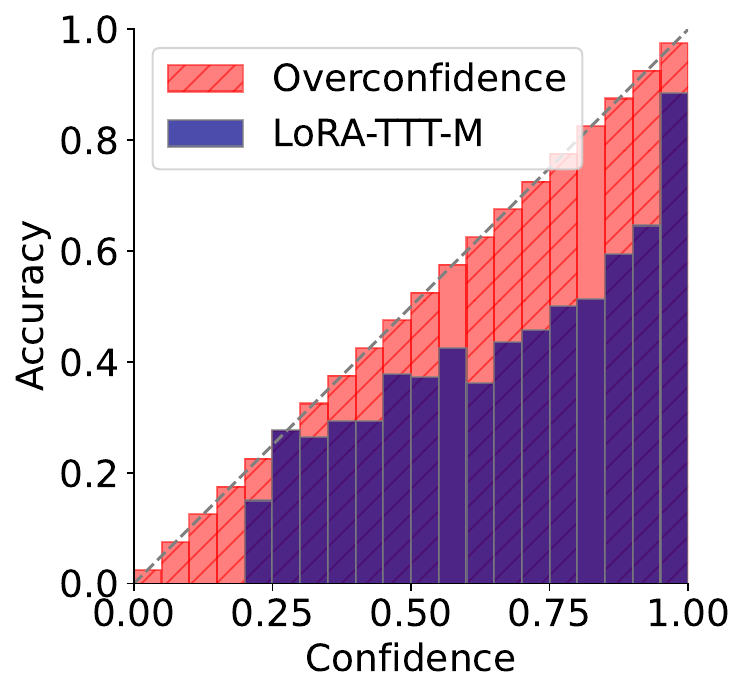}
        \subcaption{$\mathcal{L}_{\text{MEM}}$}
        \label{fig:calib_telora_m}
    \end{minipage}
    \hfill
    \begin{minipage}{0.32\linewidth}
        \centering
        \includegraphics[width=\linewidth]{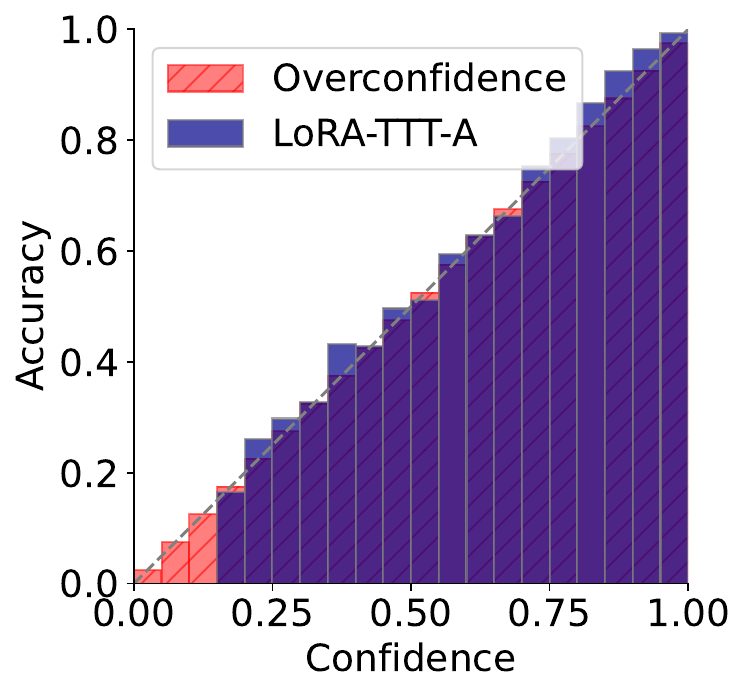}
        \subcaption{$\mathcal{L}_{\text{MAE}}$}
        \label{fig:calib_telora_a}
    \end{minipage} 
    \caption{\textbf{Comparison of calibration performance on the Cars dataset.} The MAE loss can improve performance while preserving the baseline model's output characteristics.}
    \label{fig:calib}
\end{figure}

\noindent\textbf{Calibration.}\hspace{5mm}
As shown in \cref{fig:calib_telora_m}, $\mathcal{L}_{\text{MEM}}$, commonly used as a test-time loss in VLMs, is known to induce overconfidence \citep{guo2017calibration,yoon2024c}, where the model's predicted confidence exceeds its actual accuracy.
Calibration is quantified using Expected Calibration Error (ECE) \citep{naeini2015obtaining}, which measures alignment between predicted probabilities and actual outcomes.
\cref{tab:calib} compares the ECE for our loss functions, TPT, and C-TPT.
Since \namemae\ (\ie, MAE loss) is not explicitly designed based on confidence, it retains its original calibration properties.
\namemae\ achieves ECE performance comparable to or better than C-TPT, without any specific efforts for calibration, even though C-TPT is designed to improve the calibration performance of TPT.
From a TTT perspective, the MAE loss demonstrates advantages not only in generalization across various domains but also in calibration, which is essential given the critical role of prediction uncertainty in real-world applications such as healthcare diagnostics \citep{wang2022medclip,liu2023clip} and autonomous vehicles \citep{dorbala2022clip,khandelwal2022simple}.
\newline

\begin{figure}[t]
  \centering
    \begin{minipage}{0.49\linewidth}
        \centering
        \includegraphics[width=\linewidth]{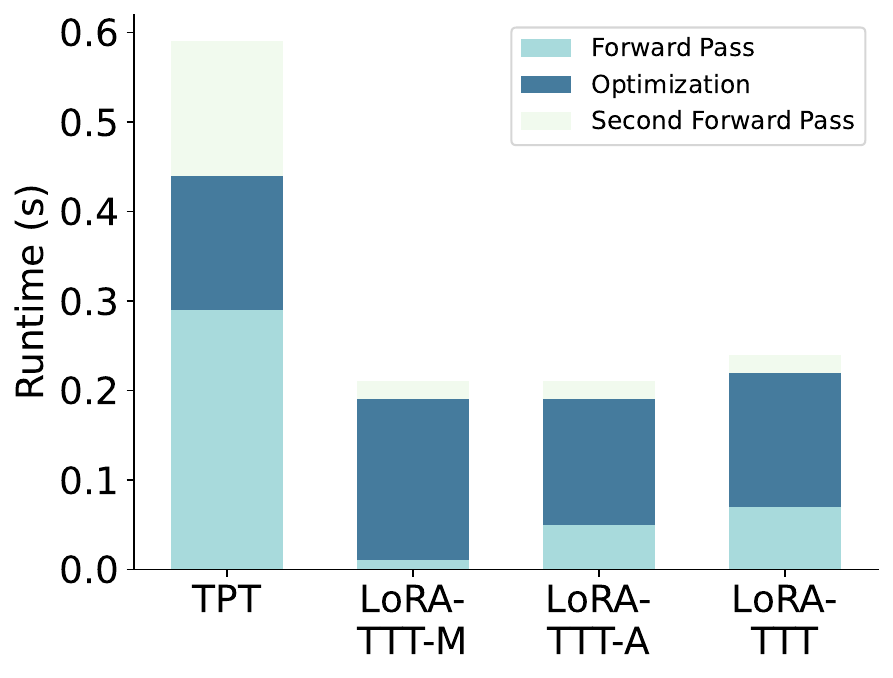}
        \subcaption{Runtime}
        \label{fig:runtime}
    \end{minipage}
    \hfill
    \begin{minipage}{0.49\linewidth}
        \centering
        \includegraphics[width=\linewidth]{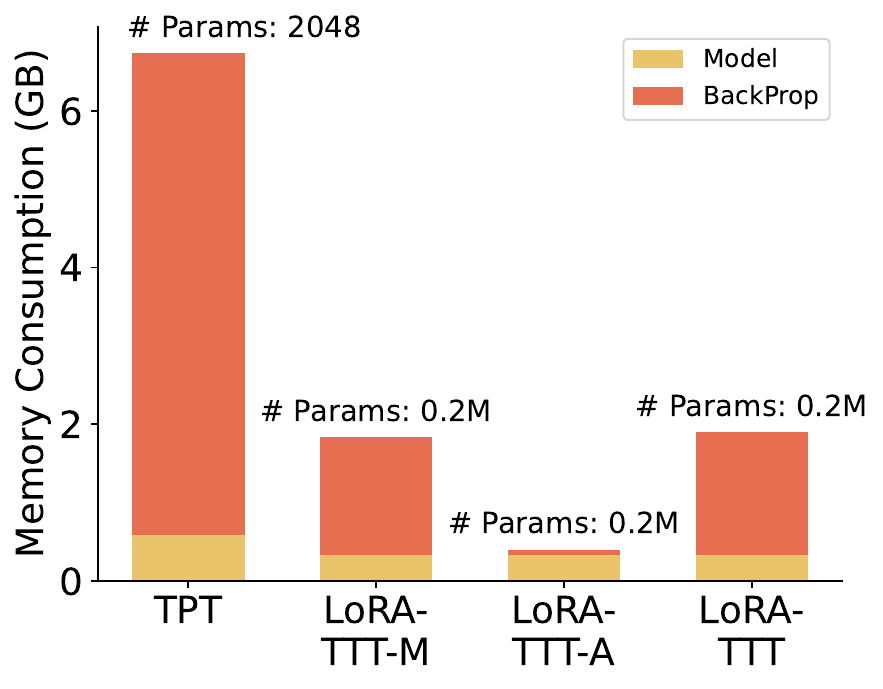}
        \subcaption{Memory Consumption}
        \label{fig:memory}
    \end{minipage}
    \caption{\textbf{TTT efficiency in ImageNet evaluation.} The efficiency of TPT heavily depends on the number of classes and input text tokens. To optimize TPT, the number of tokens is set to 20, including the 4 learnable tokens, matching the length of the longest class name in the dataset.}
    \label{fig:efficiency}
\end{figure}

\noindent\textbf{Efficiency.}\hspace{5mm} 
Runtime and memory consumption, shown in \cref{fig:runtime} and \cref{fig:memory}, demonstrate the efficiency of our approach compared to TPT.
Our method precomputes text features and eliminates the need for a text encoder during TTT.
This reduces model size and shortens forward pass times both before and after optimization compared to TPT where text encoding often serves as a bottleneck.
Although \name\ has a substantially larger number of trainable parameters than TPT, it limits LoRA application to the deeper layers of the image encoder, reducing memory usage during backpropagation.
Notably, \namemae\ significantly reduces memory usage during backpropagation, primarily by calculating the loss from only 10\% of the augmented images and masking half of the tokens.
Consequently, it requires minimal additional resources even when the MAE loss is included in the total loss.
\name\ achieves higher efficiency than TPT while delivering significantly superior performance.
Based on these results, \name\ is adaptable to a wide range of domains and applications, including streaming data processing \citep{wang2023test,azimi2022self}, from high-stakes environments \citep{wang2022medclip,liu2023clip,dorbala2022clip,khandelwal2022simple} to memory-constrained edge devices \citep{cai2020tinytl,song2023ecotta}.

\begin{table}[t]
\centering
\caption{\textbf{Expected Calibration Error} ($\downarrow$).}
\label{tab:calib}
\begin{adjustbox}{width=0.8\linewidth}
\begin{tabular}{lccc}
\toprule
\rowcolor{gray!10} \textbf{Method} & \textbf{ImageNet} & \textbf{OOD Average} & \textbf{FG Average} \\
\midrule
CLIP-ViT-B/16 & 1.93 & 4.80 & 4.53 \\
\midrule
TPT & 10.61 & 12.08 & 11.71 \\
C-TPT & 3.11 & 5.38 & 5.29 \\
\midrule
\textbf{\namemem} & 20.32 & 22.76 & 19.73 \\
\rowcolor{blue!10} \textbf{\namemae} & 2.97 & 5.59 & 4.80 \\
\textbf{\name} & 14.04 & 16.49 & 12.75 \\
\bottomrule
\end{tabular}
\end{adjustbox}
\end{table}

\begin{comment}
\noindent\textbf{Efficiency.}\hspace{5mm} 
Runtime and memory consumption, shown in \cref{fig:runtime} and \cref{fig:memory}, demonstrate the efficiency of our approach compared to TPT.
Our method precomputes text features and eliminates the need for a text encoder during TTT.
This reduces model size and shortens forward pass times both before and after optimization compared to TPT where text encoding often serves as a bottleneck.
Although \name\ has a substantially larger number of trainable parameters than TPT, it limits LoRA application to the deeper layers of the image encoder, reducing memory usage during backpropagation.
Notably, \namemae\ significantly reduces memory usage during backpropagation, primarily by calculating the loss from only 10\% of the augmented images and masking half of the tokens.
Consequently, it requires minimal additional resources even when the MAE loss is included in the total loss.
\name\ achieves higher efficiency than TPT while delivering significantly superior performance.
Based on these results, \name\ is adaptable to a wide range of domains and applications, including streaming data processing \citep{wang2023test,azimi2022self}, from high-stakes environments \citep{wang2022medclip,liu2023clip,dorbala2022clip,khandelwal2022simple} to memory-constrained edge devices \citep{cai2020tinytl,song2023ecotta}.
\end{comment}

\section{Ablation Study}
\label{sec:ablation}

\subsection{How to apply LoRA for TTT}
\label{abs:lora}
In this section, we explore the utilization of LoRA for TTT.
We investigate the key factors for effectively applying LoRA, including: (1) determining the optimal layers and the extent of LoRA application within the transformer model, (2) understanding the relationship between the appropriate rank and scale, and (3) selecting the attention matrices for tuning.
\newline

\begin{table}[t]
\centering
\caption{\textbf{Layers for LoRA application.}}
\label{tab:layer}
\begin{adjustbox}{width=0.8\linewidth}
\begin{tabular}{c|ccc}
\toprule
\rowcolor{gray!10} \textbf{LoRA Layer} & \textbf{ImageNet} & \textbf{OOD Average} & \textbf{FG Average} \\
\midrule
12 & \textbf{69.59} & 62.65 & 64.68 \\
\rowcolor{blue!10} 11-12 & 69.40 & \textbf{62.93} & \textbf{65.00} \\
9-12 & 68.97 & 62.86 & 64.73 \\
5-8 & 66.88 & 61.34 & 64.83 \\
1-4 & 67.99 & 60.69 & 64.56 \\
All & 68.12 & 62.54 & 64.62 \\
\bottomrule
\end{tabular}
\end{adjustbox}
\end{table}

\begin{figure}[t]
    \centering
    \begin{minipage}{0.49\linewidth}
        \centering
        \includegraphics[width=\linewidth]{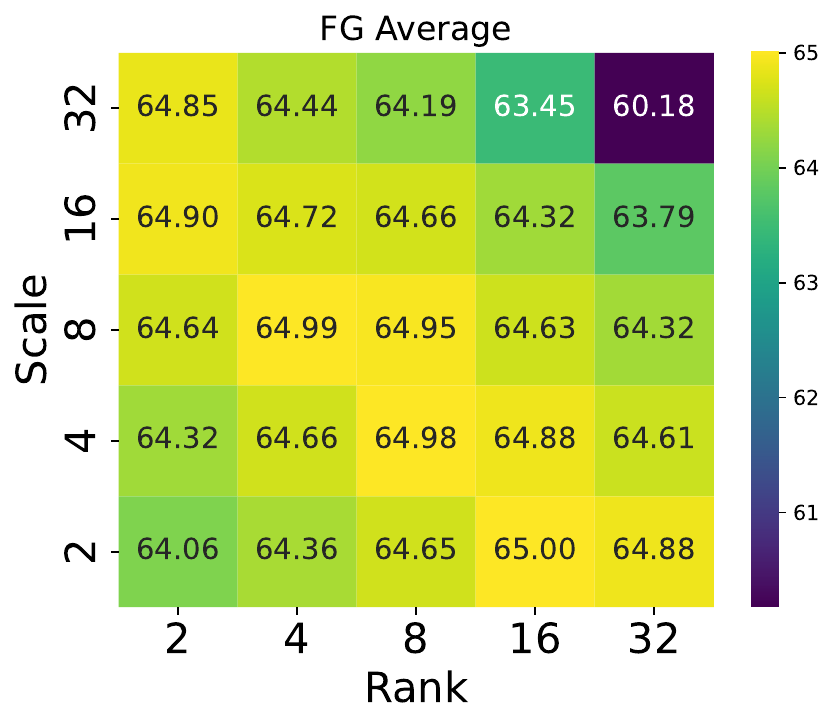}
        \subcaption{LoRA rank and scale}
        \label{fig:lora_rank_scale}
    \end{minipage}
    \hfill
    \begin{minipage}{0.49\linewidth}
        \centering
        \includegraphics[width=\linewidth]{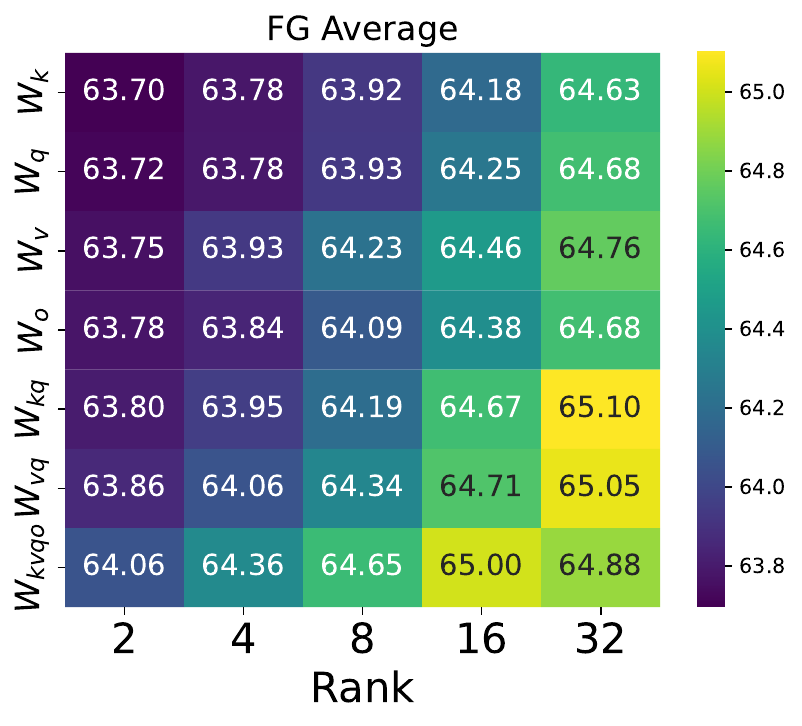}
        \subcaption{Attention matrices}
        \label{fig:lora_rank_matrix}
    \end{minipage}
    \caption{\textbf{Impact of LoRA application design.} The average top-1 accuracy on the fine-grained benchmark is shown, with LoRA applied to layers 11 and 12 of the image encoder.}
    \label{fig:lora}
\end{figure}

\noindent\textbf{Which layers should we apply LoRA to?}\hspace{5mm}
\cref{tab:layer} presents the zero-shot classification performance when LoRA is applied to specific layers of the image encoder in CLIP-ViT-B/16.
Our results indicate that applying LoRA to deeper layers is more effective than to shallower ones, aligning with trends observed in fine-tuning language models \citep{zhang2023adalora}.
Additionally, applying LoRA to more layers does not necessarily improve performance.
Limiting its application to the 11th and 12th layers not only outperforms applying it across all layers in terms of performance but also reduces memory consumption and runtime, making our approach more efficient for TTT.
\newline

\noindent\textbf{LoRA rank and scale.}\hspace{5mm}
As shown in ~\cref{fig:lora_rank_scale}, increasing the rank does not directly lead to performance gains.
Each rank has an optimal scale, and as the rank increases, the corresponding optimal scale tends to decrease.
When the rank is small (\eg, rank 4), performance remains stable across different scales, reducing the need for extensive hyperparameter tuning.
\newline

\noindent\textbf{LoRA rank and attention matrices.}\hspace{5mm}
We investigate the optimal application of LoRA to different attention matrices in CLIP-ViT-B/16.
In \cref{fig:lora_rank_matrix}, we observe that applying LoRA to $\mW_{\text{v}}$ at the same rank achieves the best results among the 4 matrices ($\mW_{\text{o}}$, $\mW_{\text{v}}$, $\mW_{\text{q}}$, and $\mW_{\text{k}}$).
This trend aligns with previous research \citep{zhang2023adalora,zanella2024low}, even in the context of TTT.
% Additionally, a trend is observed wherein $\mW_{\text{vq}}$ generally outperforms $\mW_{\text{kq}}$.
Given the same total number of parameters, applying LoRA to $\mW_{\text{kvqo}}$ shows little difference in performance compared to applying it to $\mW_{\text{vq}}$ or $\mW_{\text{kq}}$.

\subsection{Masking strategy}
\label{sec:masking}
In masked image modeling, the mask strategy plays a crucial role \citep{hondru2024masked,gao2024mimic}.
We examine the effects of the masking ratio, the confidence selection cutoff, the use of an image decoder, and the impact of reconstruction targets.
We use a randomly initialized transformer-based decoder with 8 layers, 16 heads, and a 768 embedding size, without prior fine-tuning to ensure a fair evaluation.
This decoder allows us to incorporate the pixel-wise reconstruction loss proposed in TTT methods based on MAE \citep{gandelsman2022test, wang2023test}.

As shown in \cref{tab:mask}, while the masking ratio does not significantly affect the overall performance, we choose a default masking ratio of 50\% as it strikes a good balance between performance and computational efficiency.
As proposed in TPT, selecting and masking the top 10\% of augmented images with the lowest entropy yields better performance than masking all 64 images (\ie, applying a cutoff of 1), with an improvement of over 1\% observed in the OOD average.
The 10\% cutoff not only improves performance but also enhances the computational efficiency of TTT by calculating the loss on only one-tenth of the images.
Furthermore, reconstructing the class token is more effective than reconstructing masked visual tokens or image pixels using the decoder.
This supports the hypothesis that improving zero-shot image classification performance in VLMs relies more on aligning high-level semantics than on capturing fine-grained features.
% This supports the hypothesis that reconstructing global features enhances visual understanding, thereby improving zero-shot image classification performance in VLMs.
%\citep{gao2024mimic}

\begin{table}[t]
\centering
\caption{\textbf{Masking strategy.} The LoRA scale $\gamma$ is set to 2 for both benchmarks. Performance differences from zero-shot CLIP-ViT-B/16 are shown with a blue {\textcolor{blue}{($\uparrow$)}} or red {\textcolor{red}{($\downarrow$)}} arrow.}
\label{tab:mask}
\begin{adjustbox}{width=\linewidth}
\begin{tabular}{cccc|ccc}
\toprule
\rowcolor{gray!10} \textbf{Reconstruction} & \textbf{Mask Ratio} & \textbf{Cutoff} & \textbf{Decoder} & \textbf{ImageNet} & \textbf{OOD Average} & \textbf{FG Average} \\
\midrule
\multirow{6}{*}{Class token} 
 & 0.25 & 0.1 &  & 67.65\textsubscript{\textcolor{blue}{($\uparrow$0.94)}} & \textbf{58.94}\textsubscript{\textcolor{blue}{($\uparrow$1.80)}} & 64.57\textsubscript{\textcolor{blue}{($\uparrow$0.92)}} \\
 & \cellcolor{blue!10}0.5 & \cellcolor{blue!10}0.1 & \cellcolor{blue!10} & \cellcolor{blue!10}\textbf{67.78}\textsubscript{\textcolor{blue}{($\uparrow$1.07)}} & \cellcolor{blue!10}58.85\textsubscript{\textcolor{blue}{($\uparrow$1.71)}} & \cellcolor{blue!10}\textbf{64.72}\textsubscript{\textcolor{blue}{($\uparrow$1.08)}} \\
 & 0.75 & 0.1 & & 67.48\textsubscript{\textcolor{blue}{($\uparrow$0.76)}} & 57.87\textsubscript{\textcolor{blue}{($\uparrow$0.72)}} & 64.35\textsubscript{\textcolor{blue}{($\uparrow$0.71)}} \\
 & 0.5 & 0.5 & & 67.52\textsubscript{\textcolor{blue}{($\uparrow$0.80)}} & 58.30\textsubscript{\textcolor{blue}{($\uparrow$1.16)}} & 64.49\textsubscript{\textcolor{blue}{($\uparrow$0.84)}} \\
 & 0.5 & 1 & & 67.20\textsubscript{\textcolor{blue}{($\uparrow$0.48)}} & 57.79\textsubscript{\textcolor{blue}{($\uparrow$0.65)}} & 64.34\textsubscript{\textcolor{blue}{($\uparrow$0.69)}} \\
 & 0.5 & 0.1 & \ding{51} & 67.27\textsubscript{\textcolor{blue}{($\uparrow$0.55)}} & 58.28\textsubscript{\textcolor{blue}{($\uparrow$1.14)}} & 64.14\textsubscript{\textcolor{blue}{($\uparrow$0.50)}} \\
\midrule
Visual tokens & 0.5 & 0.1 & & 66.89\textsubscript{\textcolor{blue}{($\uparrow$0.17)}} & 57.46\textsubscript{\textcolor{blue}{($\uparrow$0.32)}} & 63.79\textsubscript{\textcolor{blue}{($\uparrow$0.15)}} \\
\midrule
Image pixel & 0.5 & 0.1 & \ding{51} & 66.67\textsubscript{\textcolor{red}{($\downarrow$0.05)}} & 57.05\textsubscript{\textcolor{red}{($\downarrow$0.10)}} & 63.50\textsubscript{\textcolor{red}{($\downarrow$0.14)}} \\
\bottomrule
\end{tabular}
\end{adjustbox}
\end{table}

\subsection{Initialization of LoRA weights}
LoRA demonstrates high effectiveness and efficiency for TTT, even when initialized with random weights.
In this section, we explore the performance gains achieved by fine-tuning the LoRA weights before TTT.
We prepare a third dataset, CC3M \citep{sharma2018conceptual}, for LoRA initialization and train only the LoRA weights using the same contrastive loss as in CLIP pre-training \citep{radford2021learning} with image-text pairs.
We employ Adam with a learning rate of $1\text{e-}6$ and a weight decay of 0.05 for optimization, performing one epoch of training with a batch size of 64.

\begin{figure}[t]
  \centering
    \begin{minipage}{0.49\linewidth}
        \centering
        \includegraphics[width=\linewidth]{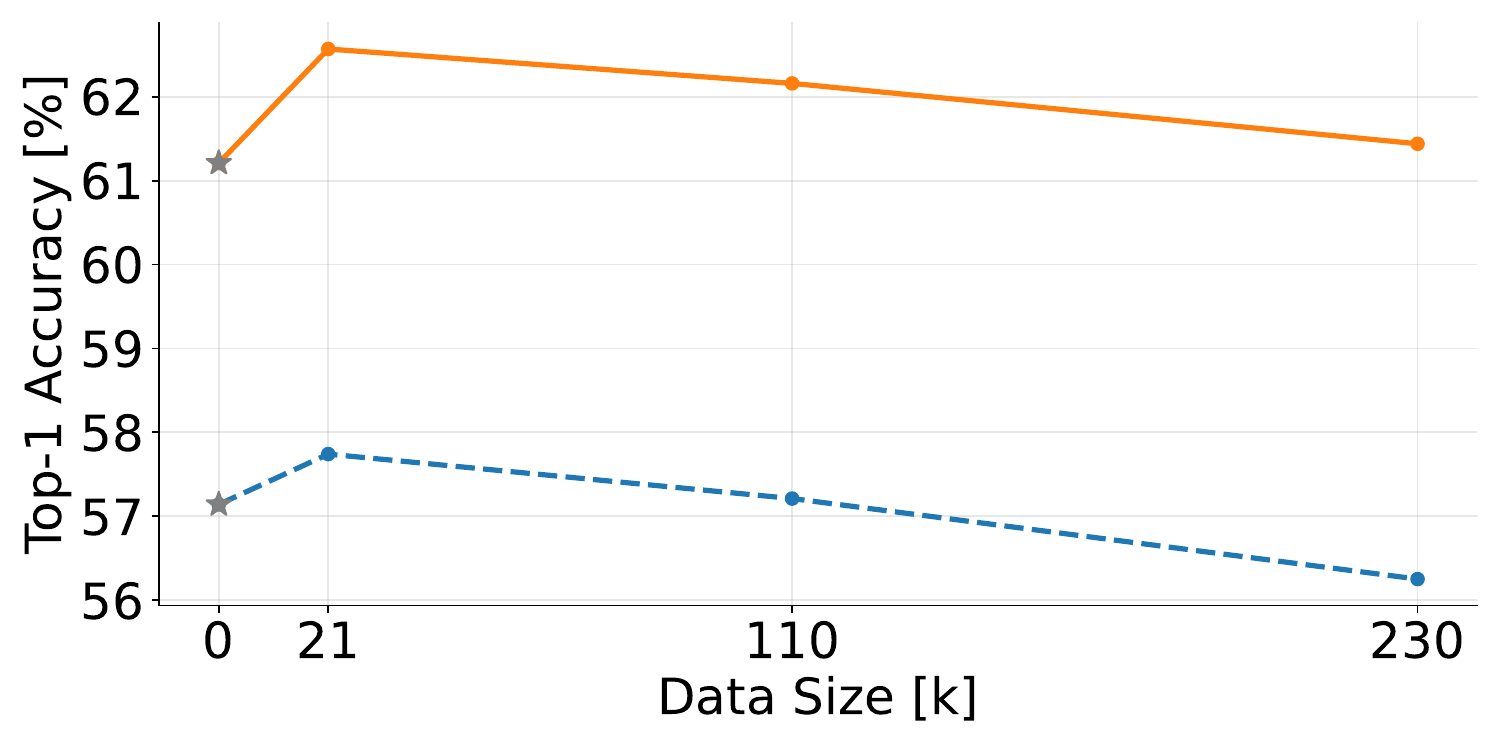}
        \subcaption{OOD Average}
        \label{fig:init_ood}
    \end{minipage}
    \hfill
    \begin{minipage}{0.49\linewidth}
        \centering
        \includegraphics[width=\linewidth]{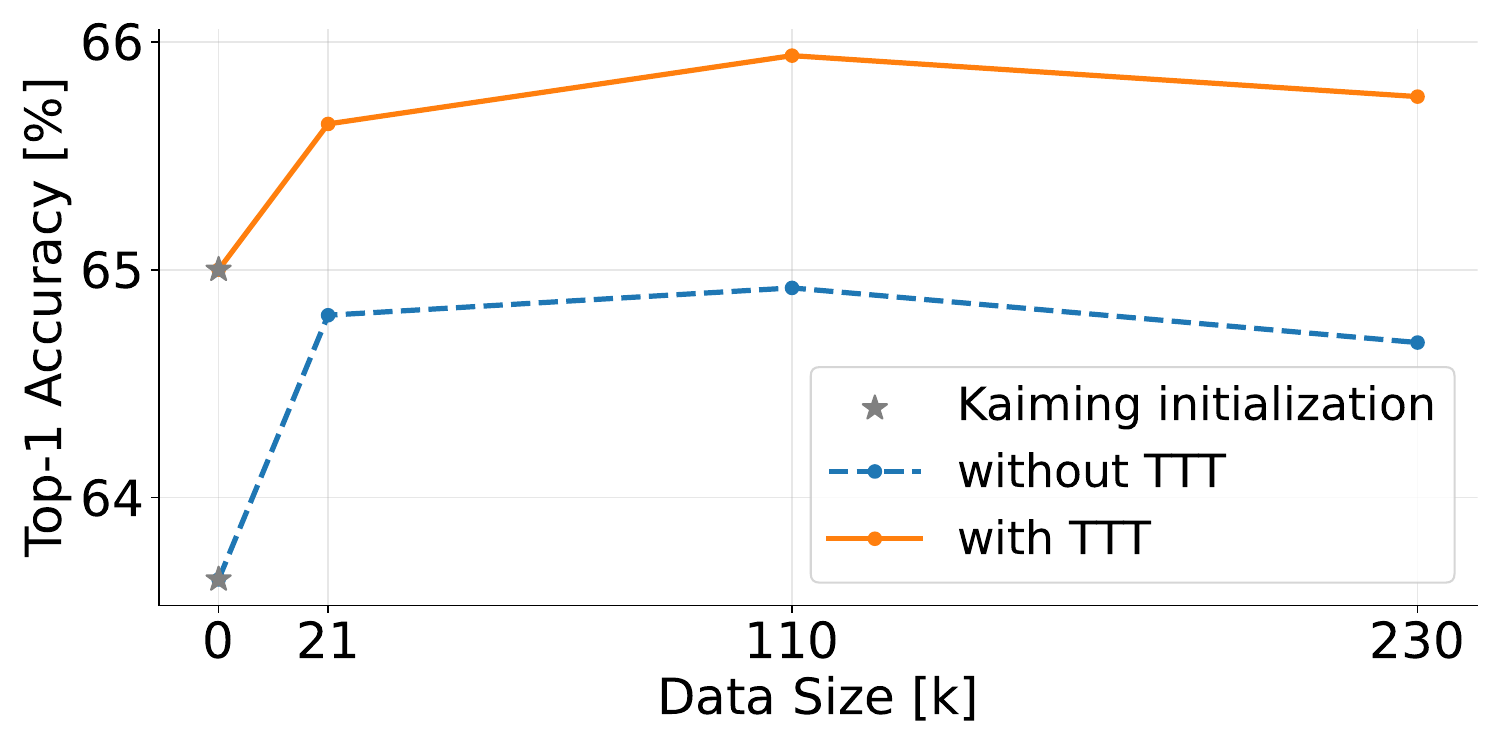}
        \subcaption{Fine-Graind Average}
        \label{fig:init_fg}
    \end{minipage}
    \caption{\textbf{Impact of LoRA weight initialization} by data size and comparison with TTT.}
    \label{fig:init}
\end{figure}

As shown in \cref{fig:init}, LoRA initialization using 21k randomly sampled image-text pairs from CC3M (\ie, only 1\% of the total CC3M dataset) improves performance by more than 1\% on the fine-grained benchmark and by 0.6\% on the OOD benchmark.
Furthermore, TTT consistently improves performance on both the benchmarks, regardless of the LoRA initialization.
% On the OOD benchmark, it exceeds the Kaiming initialization by more than 1\%, demonstrating the effectiveness of LoRA fine-tuning. 
Our experiments demonstrate that fine-tuning LoRA with a small amount of data shows the potential to enhance its performance.
While adhering to the constraints of not leveraging domain-specific information or a teacher model, LoRA fine-tuning delivers significant performance improvements in TTT, establishing it as an effective approach for future applications of LoRA in TTT.

\begin{comment}
\begin{table}[t]
\centering
\caption{\textbf{LoRA weight initialization}}
\label{tab:init}
\begin{adjustbox}{width=0.8\linewidth}
\begin{tabular}{ccc|ccc}
\toprule
Initialization & Data Size & TTT & ImageNet & OOD Average & FG Average \\
\midrule
Kaiming & 0 & & 66.72 & 57.14 & 63.64 \\
Kaiming & 0 & \ding{51} & 69.31 & 61.21 & 65.00 \\
\midrule
CC3M & 21 $\mathrm{k}$ & & 67.92 & 57.74 & 64.80 \\
CC3M & 21 $\mathrm{k}$ & \ding{51} & 70.52 & 62.57 & 65.64 \\
CC3M & 110 $\mathrm{k}$ & & 67.56 & 57.21 & 64.92 \\
CC3M & 110 $\mathrm{k}$ & \ding{51} & 70.24 & 62.16 & 65.94 \\
CC3M & 230 $\mathrm{k}$ & & 66.96 & 56.25 & 64.68 \\
CC3M & 230 $\mathrm{k}$ & \ding{51} & 69.78 & 61.44 & 65.76 \\
\bottomrule
\end{tabular}
\end{adjustbox}
\end{table}
\end{comment}

\section{Conclusion}
\label{sec:conclusion}

This paper presents a \textbf{Lo}w-\textbf{Ra}nk \textbf{T}est-\textbf{T}ime \textbf{T}raining (\name), a novel test-time training method for VLMs.
\name\ leverages LoRA, enabling effective adaptation to distribution shifts during test time without incurring catastrophic forgetting.
Additionally, we introduce a highly efficient reconstruction loss suited for TTT, which enhances the generalization and calibration performance of our method.
Extensive experiments on two benchmarks demonstrate that \name\ outperforms state-of-the-art text prompt tuning methods, while requiring less memory, runtime, and avoiding the need for external resources.
These results show that \name\ can be applied across a wide range of domains and applications, from high-stakes environments to edge devices.
We hope our work serves as a foundation for developing new TTT methods for foundation models, unlocking the potential of the image encoder.

% \section{acknowledgement}
% This research is supported by funding from Sony Semiconductor Solutions Corporation through the Visiting Industry Fellow program at the University of California, San Diego.

{
    \small
    \bibliographystyle{ieeenat_fullname}
    \bibliography{main}
}

% \clearpage
% \appendix
% \maketitlesupplementary

% WARNING: do not forget to delete the supplementary pages from your submission
% \input{sec/X_suppl}

%{
%    \small
%    \bibliographystyle{ieeenat_fullname}
%    \bibliography{main}
%}

\end{document}